\documentclass[letterpaper, 10 pt, conference]{ieeeconf}  

\IEEEoverridecommandlockouts                              

\usepackage{graphicx}
\usepackage{times}
\usepackage{epsfig}
\usepackage{float}
\usepackage{subfigure}
\usepackage{multirow}
\usepackage{amsmath,amssymb} 
\usepackage[pagebackref=true,breaklinks=true,letterpaper=true,colorlinks,bookmarks=false]{hyperref}
\usepackage{color}

\title{Motion Segmentation via Global and Local Sparse Subspace Optimization} 

\author{Michael Ying Yang$^{1}$,  Hanno Ackermann$^{2}$, Weiyao Lin$^{3}$, Sitong Feng$^{2}$ and Bodo Rosenhahn$^{2}$
\thanks{$^{1}$University of Twente, \textbf{michael.yang@utwente.nl}}%
\thanks{$^{3}$Leibniz University Hannover}%
\thanks{$^{2}$Shanghai Jiao Tong University}%
}

\begin{document}
\maketitle
\thispagestyle{empty}
\pagestyle{empty}

\begin{abstract}
   In this paper, we propose a new framework for segmenting feature-based moving objects under affine subspace model. Since the feature trajectories in practice are high-dimensional and contain a lot of noise, we firstly apply the sparse PCA to represent the original trajectories with a low-dimensional global subspace, which consists of the orthogonal sparse principal vectors. Subsequently, the local subspace separation will be achieved via automatically searching the sparse representation of the nearest neighbors for each projected data. In order to refine the local subspace estimation result and deal with the missing data problem, we propose an error estimation to encourage the projected data that span a same local subspace to be clustered together. In the end, the segmentation of different motions is achieved through the spectral clustering on an affinity matrix, which is constructed with both the error estimation and sparse neighbors optimization. We test our method extensively and compare it with state-of-the-art methods on the Hopkins 155 dataset and Freiburg-Berkeley Motion Segmentation dataset. The results show that our method is comparable with the other motion segmentation methods, and in many cases exceed them in terms of precision and computation time.
\end{abstract}

\section{Introduction}
\begin{figure*}[!htbp]
 \centering
   \includegraphics[scale=0.3]{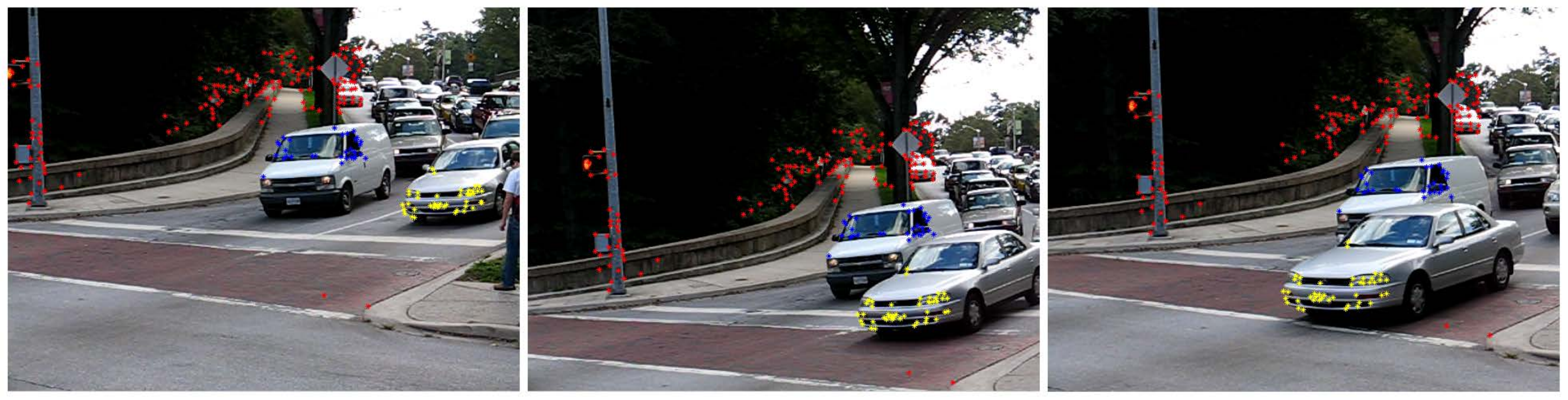}
   \caption{Example results of the motion segmentation on the real traffic video \textit{cars9.avi} from the Hopkins 155 dataset \cite{hopkins155}.}
   \label{fig:moseg_demo}
 \end{figure*}

In the past years, dynamic scenes understanding has been receiving increasing attention especially on the moving camera or multiple moving objects. Motion segmentation as a part of the video segmentation is an essential part for studying the dynamic scenes and many other computer vision applications \cite{YanRos2014a}. Particularly, motion segmentation aims to decompose a video into different regions according to different moving objects that tracked throughout the video. 
In case of feature extraction for all the moving objects from the video, segmentation of different motions is equivalent to segment the extracted feature trajectories into different clusters. One example of feature-based motion segmentation is presented in Figure~\ref{fig:moseg_demo}.

Generally, the algorithms of motion segmentation are classified into 2 categories \cite{MSMC}: affinity-based methods and subspace-based methods.
The affinity-based methods focus on computing the correspondences of each pair of the trajectories, whereas the subspace-based approaches use multiple subspaces to model the multiple moving objects in the video and the segmentation of different motions is accomplished through subspace clustering.
Recently, some affinity-based methods \cite{MSMC,PAMI} are proposed to cluster the trajectories with unlimited number of missing data. However, the computation times of them are so high that require an optimizing platform to be reduced. 
Whereas, the subspace-based methods \cite{SSC,ALC} have been developed to reconstruct the missing trajectories with their sparse representation. The drawback is that they are sensitive to the real video which contains a large number of missing trajectories. Most of the existing subspace-based methods still fall their robustness for handling missing features. Thus, there is an intense demand to explore a new subspace-base algorithm that can not only segment multiple kinds of motions but also handle the missing and corrupted trajectories from the real video.

\subsection{Contributions}
We propose a new framework with subspace models for segmenting different types of moving objects from a video under affine camera. 
We cast the motion segmentation as a two stage subspace estimation: the global and local subspace estimation. 
Sparse PCA~\cite{spca} is adopted for optimizing the global subspace in order to defend the noise and outliers. 
Meanwhile, we seek a sparse representation for the nearest neighbors in the global subspace for each data point that span a same local subspace. In order to solve the missing data problem and refine the local subspace estimation, we propose an error estimation and build the affinity graph for spectral clustering to obtain the clusters.
To the best of our knowledge, our framework is the first one to simultaneously optimize the global and local subspace with sparse representation.

\vspace{5mm}
The remaining sections are organized as follows. 
The related works are discussed in Section~\ref{sec:related}. 
The basic subspace models for motion segmentation are introduced in Section~\ref{sec:subspace}. 
The proposed approach will be described in detail in Section~\ref{sec:sparse}. 
Furthermore, the experimental results are presented in Section~\ref{sec:exp}. 
Finally, this paper is concluded in Section~\ref{sec:con}.

\section{Related Work}
\label{sec:related}

During the last decades, either the subspace-based techniques \cite{SSC,ALC} or the affinity-based methods \cite{MSMC,PAMI} have been receiving an increasing interest on segmentation of different types of motions from a real video. 

\textbf{Affinity-based methods.}
\cite{PAMI} uses the distances of each pair of feature trajectories as the measurement to build the affinity matrix based on a translational motion model. This method can segment motions with unlimited number of missing or incomplete trajectories, which means they are robust to the video with occlusions or moving camera problems.
Another approach which is based on the affinity is called Multi-scale Clustering for Motion Segmentation (MSMC) \cite{MSMC}. Based on the conventional image classification technique split and merge, they use the correspondences of each two features between two frames to segment the different motions with many missing data. 
One of the general problems of affinity-based method is highly time-consuming. They have to be implemented with an optimized platform in order to save the computation times.

\textbf{Subspace-based methods.}
The existing works based on subspace models can be divided into 4 main categories: algebraic, iterative, sparse representation and subspace estimation. 

Algebraic approaches, such as Generalized Principal Component Analysis (GPCA) \cite{GPCA}, which uses the polynomials fitting and differentiation to obtain the clusters. GPCA can segment the rigid and non-rigid motions effectively, but once the number of moving objects in the video increases, its computation cost increases and the precision decreases in the same time. 
The general procedure of an iterative method contains two main aspects: find the initial solution and refine the clustering results to fit each subspace model. RANdom SAmple Consensus (RANSAC) \cite{ransac} selects randomly the number of points from the original dataset to fit the model. RANSAC is robust to the outliers and noise, but it requires a good initial parameter selection.
 Specifically, it computes the residual of each point to the model with setting a threshold, if the residual is below the threshold, it will be considered as inliers and vice versa.  
Sparse Subspace Clustering (SSC) \cite{SSC} is one of the most popular method based on the sparse representation. SSC exploits a fact that each point can be linearly represented with a sparse combination of the rest of other data points. SSC has one of the best accuracy compared with the other subspace-based methods and can deal with the missing data. The limitation is that it requires a lot of computation times. 
Another popular algorithm based on the sparse representation is Agglomerate Lossy Compression (ALC) \cite{ALC}, which uses compressive sensing on the subspace model to segment the video with missing or corrupted trajectories. However, the implementation of ALC cannot ensure that find the global maximum with the greedy algorithm. By the way ALC is highly time-consuming in order to tune the parameter.

Our work combines the subspace estimation and sparse representation methods.
The subspace estimation algorithms, such as Local Subspace Affinity (LSA) \cite{LSA}, firstly project original data set with a global subspace. Then the projected global subspace is separated into multiple local subspaces through K-nearest neighbors (KNN). 
After calculating the affinities of different estimated local subspaces with principle angles, the final clusters are obtained through spectral clustering. 
It comes to the issue that the KNN policy may overestimate the local subspaces due to noise and improper selection of the number K, which is determined by the rank of the local subspace. 
LSA uses the model selection (MS) \cite{modelselection} to estimate the rank of global and local subspaces, but the MS is quite sensitive to the noise level.

\section{Multi-body Motion Segmentation with Subspace models}
\label{sec:subspace}

In this section, we introduce the motion structure under affine camera model. Subsequently, we show that under affine model segmentation of different motions is equivalent to separate multiple low-dimensional affine subspaces from a high-dimensional space.

\subsection{Affine Camera Model}
Most of the popular algorithms assume an affine camera model, which is an orthographic camera model and has a simple mathematical form. It gives us a tractable representation of motion structure in the dynamic scenes. 
Under the affine camera, the general procedure for motion segmentation is started from translating the 3-D coordinates of each moving object to its 2-D locations in each frame. 
Assume that $\{x_{fp}\}_{f=1, ..., F}^{p=1, ..., P}\in R^2$ represents one 2-D tracked feature point $p$ of one moving object at frame $f$, its corresponding 3-D world coordinate is $\{ X_p \}_{p=1, ..., P} \in R^3$. The pose of the moving object at frame $f$ can be represented with $(R_f, T_f) \in SO(3)$, where $R_f$ and $T_f$ are related to the rotation and translation respectively. Thus, each 2-D point $x_{fp}$ can be described with Equation \ref{eq:affine}
\begin{equation}
x_{fp} = \left[ R_f \ T_f \right] X_p = A_f X_p
\label{eq:affine}
\end{equation}
where $A_f=\left[\begin{array}{ccc}
1 & 0 & 0\\
0 & 1 & 0\\
\end{array}\right] \left[R_f \ T_f \right]\in R^{2\times 4}$ is the affine transformation matrix at frame $f$. 
\subsection{Subspace models for Motion Segmentation under Affine View}

The general input for the subspace-based motion segmentation under affine camera can be formulated as a trajectory matrix containing the 2-D positions of all the feature trajectories tracked throughout all the frames. 
Given 2-D locations $\{x_{fp}\}_{f=1, ..., F}^{p=1, ..., P}\in R^2$ of the tracked features on a rigid moving object, the corresponding trajectory matrix can be formulated as Equation \ref{eq:trajectoryMatrix}
\begin{equation}
W_{2F\times P}= \left[\begin{array}{ccc}
x_{11} & \cdots & x_{1P} \\
 \vdots & \vdots & \vdots \\
x_{F1} & \cdots & x_{FP} \\
\end{array}\right]
\label{eq:trajectoryMatrix}
\end{equation}
under affine model, the trajectory matrix $W_{2F\times P}$ can be further reformulated as Equation \ref{eq:affineW}
\begin{equation}
W_{2F\times P}= \left[\begin{array}{c}
A_1 \\
\vdots \\
A_F \\
\end{array}\right]_{2F\times 4} \left[\begin{array}{ccc}
X_1 & \cdots & X_P\\
1 & \cdots & 1 \\
\end{array} \right]_{4\times P}
\label{eq:affineW}
\end{equation}
we can rewrite it as following,
\begin{equation}
W_{2F\times P}=M_{2F\times 4}S_{P\times 4}^T
\label{affineSimple}
\end{equation}
where $M_{2F\times 4}$ is called motion matrix, whereas $S_{P\times 4}$ is structure matrix. According to Equation \ref{affineSimple}, we can obtain that under affine view the rank of trajectory matrix $W_{2F\times P}$ of a rigid motion is no more than 4. 
Hence, as the trajectory matrix is obtained, the first step is reducing its dimensionality with a low-dimension representation, which is called the global subspace transformation. 
Subsequently, each projected trajectory from the global subspace lives in a local subspace. 
Then the obstacle of multi-body motion segmentation is to separate these underlying local subspaces from the global subspace, which means the segmentation of different motions is related with segmenting different subspaces.

\section{Proposed Framework}
\label{sec:sparse}

\begin{figure*}[!htbp]
\begin{center}
\includegraphics[width=1.09\textwidth, height=0.36\textheight]{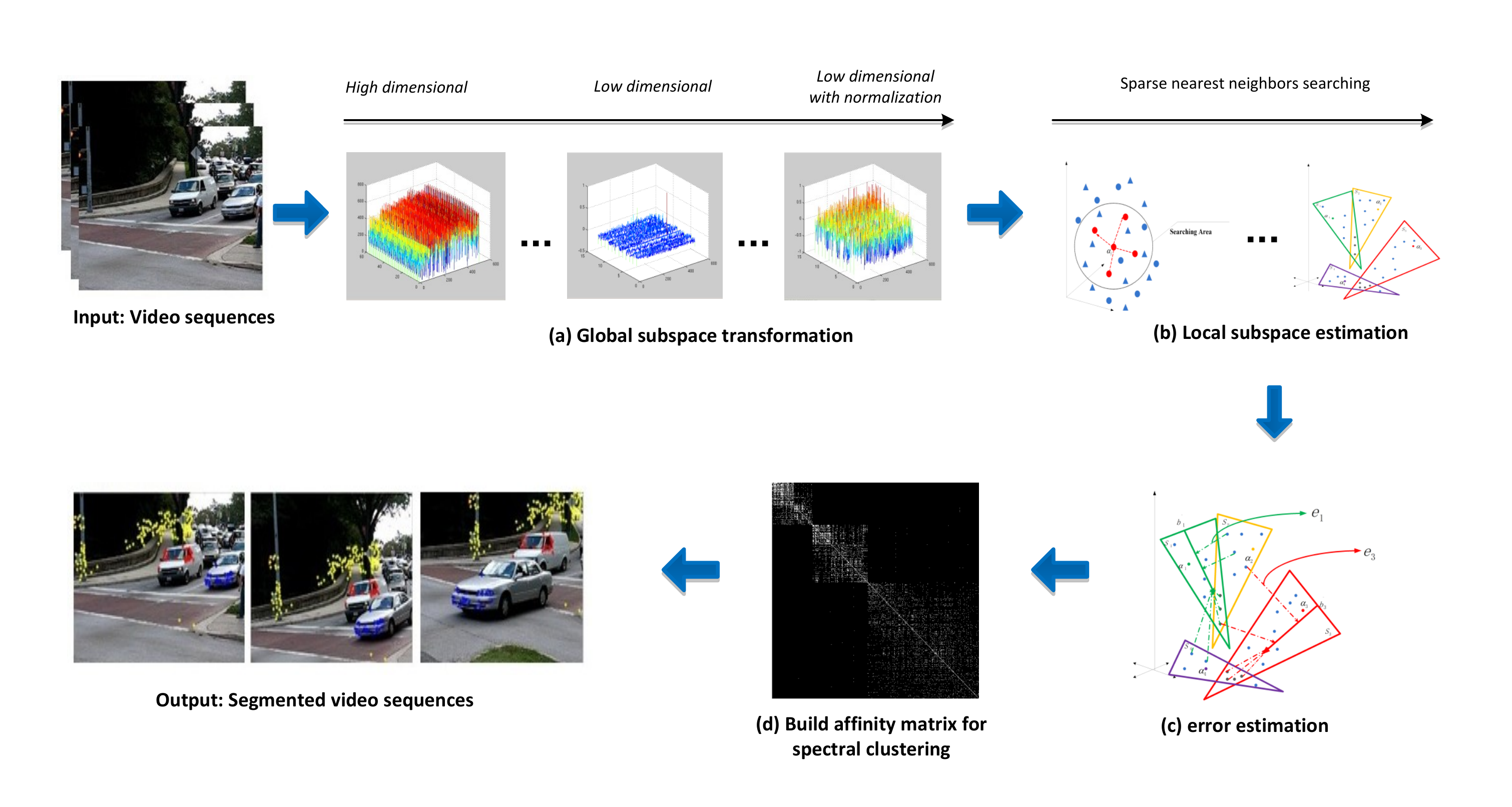}
\end{center}
   \caption{Overview of the proposed framework.}
\label{fig:overview}
\end{figure*}

Our proposed framework extends the LSA \cite{LSA} with sparse optimization for both the global and local parts. As shown in Figure~\ref{fig:overview}, given a general trajectory matrix, we firstly transform it into a global subspace with Sparse PCA \cite{spca}, which is robust to noise and outliers. 
Furthermore, instead of the KNN estimation we use the sparse neighbors to automatically find the projected data points span a same subspace. 
To correct the overestimation and encourage the projected data from the same subspace to be collected, we propose an error function to build the affinity matrix for spectral clustering.
%

\subsection{Global Subspace Transformation}
Due to the trajectory matrix of a rigid motion has a maximal rank 4, most people choose the projected dimension to be $m=4n$ or $5$, where $n$ is the number of the motions in the video. 
Assume that the trajectory matrix is $W_{2F\times P}$, where $F$ is the number of frames and $P$ is the number of extracted trajectories.
The traditional way to project $W_{2F\times P}$ is Principal Component Analysis (PCA) \cite{pca}, which can be formed as following,
\begin{equation}
z^*=\max_{z^Tz \leq 1}z^T\Sigma z,
\label{eq:pca}
\end{equation}
where $\Sigma=W^TW$ is the covariance matrix of $W$, solutions $z^*$ represent the principal components. Usually, PCA can be obtained through performing singular value decomposition (SVD) for $W$. The solutions $z^*$ are fully observed, which means they are constructed with all the input variables. However, if the principal components $z^*$ are built with only a few number of original variables but still can represent the original data matrix well, it should be easier to separate the underlying local subspaces from the transformed global subspace.
The sparse PCA technique has been proved that it is robust to the noise and outliers in terms of dimensionality reduction and feature selection \cite{SPCA_dimensionReduction,SPCA_featureselection}, which aims to seek a low-dimensional sparse representation for the original high-dimensional data matrix.
In contrast to PCA, sparse PCA produces the sparse principal components that achieve the dimensional reduction with a small number of input variables but can interpret the main structure and significant information of the original data matrix. 

In order to contain the orthogonality of projected vectors in the global subspace, we apply the generalized power method for sparse PCA \cite{gspca} to transform the global subspace.
Given the trajectory matrix $W_{2F\times P}=\left[w_1, ..., w_{F} \right]^T$, where $w_f \in R^{2\times P}, f=1, ...,F$ contains all the tracked $P$ 2-D feature points in each frame $f$.  
We can consider a direct single unit form as Equation \ref{eq:one_spca} to extract one sparse principal component $z^*\in R^P$ \cite{spca,gspca}.  
\begin{equation}
z^*(\gamma )=\max\limits_{y\in B^P}\max\limits_{z\in B^{2F}}(y^TWz)^2-\gamma \| z \|_0
\label{eq:one_spca}
\end{equation} 
where $y$ denotes a initial fixed data point from the unit Euclidean sphere $B^P=\{y\in R^P|y^Ty \leq 1\}$, and $\gamma>0$ is the sparsity controlling parameter. 
If project dimension is $m, 1<m<2F$, which means there are more than one sparse principal components needed to be extracted, in order to enforce the orthogonality for the projected principal vectors, \cite{gspca} extends Equation \ref{eq:one_spca} to block form with a trace function(\textit{Tr()}), which can be defined as Equation \ref{eq:multi_spca}
\begin{equation}
\begin{aligned}
Z^*(\gamma)=&\max\limits_{Y \in S_m^P}\max\limits_{Z\in [S^{2F}]^m}Tr(Diag(Y^TWZN)^2) \\
            &-\sum_{j=1}^m\gamma_j \|z_j\|_0
\end{aligned}
\label{eq:multi_spca}
\end{equation}
where $\gamma = \left[\gamma_1, ..., \gamma_m \right]^T$ is a positive $m$-dimensional sparsity controlling parameter vector, and parameter matrix $N = Diag(\mu_1, \mu_2, ..., \mu_m)$ with setting distinct positive diagonal elements enforces the loading vectors $Z^*$ to be more orthogonal, 
$S_m^p=\{Y \in R^{P\times m}|Y^TY=I_m \}$ represents the \textit{Stiefel manifold}\footnote {Stiefel manifold: the Stiefel manifold $V^k(R^n)$ is the set of all orthonormal k-frames in $R^n$.}.
Subsequently, Equation \ref{eq:multi_spca} is completely decoupled in the columns of $Z^*(\gamma)$ as following,
\begin{equation}
Z^*(\gamma)=\max_{Y\in S_m^P}\sum_{j=1}^m\max_{z_j\in S^{2F}}(\mu_jy_j^TWz_j)^2-\gamma_j||z_j||_0
\label{eq:multi_spca2}
\end{equation}
Obviously, the objective function in Equation \ref{eq:multi_spca2} is not convex, but the solution $Z^*{\gamma}$ can be obtained after solving a convex problem in Equation \ref{eq:multi_spcaconvex}
\begin{equation}
Y^*(\gamma)=\max\limits_{Y\in S_m^P}\sum_{j=1}^m\sum_{i=1}^{F}\left[(\mu_j w_i^T y_j)^2-\gamma_j \right]_+
\label{eq:multi_spcaconvex}
\end{equation}
which under the constraint that all $\gamma_j>\mu_j^2\max_i||w_i||_2^2$. 
In \cite{gspca}, a gradient scheme has been proposed to efficiently solve the convex problem in Equation \ref{eq:multi_spcaconvex}. 
Hence, the sparsity pattern $\mathbf{I}$ for the solution $Z^*$ is defined by $Y^*$ after Equation \ref{eq:multi_spcaconvex} under the following criterion,
\begin{equation}
\mathbf{I} = \left\{ \begin{array}{cc}
active, & (\mu_jw_i^Ty_j^*)^2>\gamma_j,\\
0, & otherwise\\
\end{array}\right.
\end{equation}
As a result, the seeking sparse loading vectors $Z^*\in S_m^P$ are obtained after iteratively solving Equation \ref{eq:multi_spcaconvex}. After normalization, the global projected subspace $\widetilde{W}_{m\times P}=normalize(Z^*)^T $ is achieved, which is embedded with multiple orthogonal underlying local subspaces.

\subsection{Local Subspace Estimation}

\begin{figure*}
\begin{center}
\includegraphics[scale=0.33]{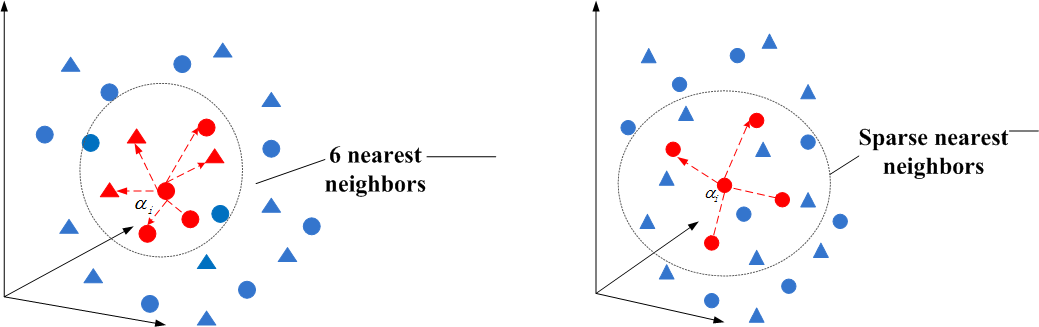}
\end{center}
   \caption{Illustration of 6-nearest neighbors and sparse nearest neighbors policy. The circles and triangles represent the data points from two different local subspaces respectively. The red points denote the estimated neighbors for the observed data $\alpha_i$ from the same local subspace under the determinate searching area.}
\label{fig:sparse_neighbors}
\end{figure*}

In order to cluster the different subspaces according to different moving bodies, the first step is finding out the multiple underlying local subspaces from the global subspace. Generally, the estimation of different local subspaces can be addressed as the extraction of different data sets, which contain only the projected trajectories from the same subspace. One of the most traditional way is the local sampling \cite{LSA}, which uses the KNN. Specifically, the underlying local subspace spanned by each projected data is found by collecting each projected data point and its corresponding K nearest neighbors, which are calculated by the distances \cite{LSA,LLMC}. 
However, the local sampling can not ensure that all the extracted K-nearest neighbors truly span one same subspace, which means an overestimation, especially for the video who contains a lot of degenerated/depended motions or missing data. Moreover, \cite{ELSA} has testified that the selection of number K is quite sensitive, which depends on the rank estimation.
In this paper, for the sake of avoiding the searching for only nearest neighbors and solving the overestimation problem we adopt the sparse nearest neighbors optimization to automatically find the set of the projected data points that span a same local subspace.

The assumption of sparse nearest neighbors is derived from SMCE \cite{SMCE}, which can cluster the data point from a same manifold robustly. 
Given a random data point $x_i$ that draw from a manifold $M_l$ with dimension $d_l$, under the SMCE assumption, we can find a relative set of points $\mathcal{N}_i={x_j, j\neq i}$ from $M_l$ but contains only a small number of non-zero elements that passes through $x_i$. This assumption can be mathematically defined with Equation \ref{eq:smce}
\begin{equation}
\|c_i[x_1-x_i, ..., x_P-x_i]\|_2 \leq \epsilon, \ s.t \ \textbf{1}^Tc_i=\textbf{1} 
\label{eq:smce}
\end{equation}
where $c_i$ contains only a few non-zero entries that denote the indices of the data point that are the sparse neighbors of $x_i$ from the same manifold, $\textbf{1}^Tc_i=\textbf{1}$ is the affine constraint and $P$ represent the number of all the points lie in the entire manifold. 

We apply the sparse neighbors estimation to find the underlying local subspaces in our transformed global subspace. As shown in Figure \ref{fig:sparse_neighbors}, with the 6-nearest neighbors estimation, there are four triangles have been selected to span the same local subspace with observed data $\alpha_i$, because they are near to $\alpha_i$ than the other circles. While, the sparse neighbors estimation is looking for only a small number of data point that close to $\alpha_i$, in this way most of the intersection area between the different local subspaces can be eliminated.
In particular, we constraint the searching area of the sparse neighbors for each projected trajectory from the global subspace with calculating the normalized subspace inclusion (NSI) distances \cite{NSI} between them. NSI can give us a robust measurement between the orthogonal projected vectors based on their geometrically consistent, which is formulated as
\begin{equation}
NSI_{ij}=\frac{tr\{\alpha_i^T \alpha_j \alpha_j^T \alpha_i \}}{\min(\dim(\alpha_i), \dim(\alpha_j))}
\label{eq: nsi_distance}
\end{equation}
where the input is the projected trajectory matrix $\widetilde{W}_{m\times P}=[\alpha_1, ..., \alpha_P]$, and $\alpha_i$ and $\alpha_j, i,j=1, ..., P$ represent two different projected data. The reason of using NSI distances to constraint the sparse neighbors searching area is the geometric property of the projected global subspace. Nevertheless the data vectors which are very far away from $\alpha_i$ definitely can not span the same local subspace with $\alpha_i$. Moreover, in addition to save computation times, the selection for the searching area with NSI distances is more flexible, which has a wide range of values, than tuning the fixed parameter K for nearest neighbors.

Furthermore, all the NSI distances are stacked into a vector $X_i=[NSI_{i1}, ..., NSI_{iP}]^T$, the assumption from SMCE in Equation \ref{eq:smce} can be solved with a weighted sparse  $\mathcal{L}_1$ optimization under affine constraint, which is formulated as following
\begin{equation}
\begin{aligned}
&\min\|Q_ic_i\|_1 \\
&s.t \ \|X_ic_i\|_2 \leq \epsilon, 1^Tc_i=1
\end{aligned}
\label{eq:solve_sparseneighbor}
\end{equation}
where $Q_i$ is a diagonal weight matrix and defined as $Q_i=\frac{exp(X_i/\sigma)}{exp(\sum_t\neq i X_{it})/\sigma}\in (0, 1], \sigma > 0$. The effect of the positive-definite matrix $Q_i$ is encouraging the selection of the closest points for the projected data $\alpha_i$ with a small weight, which means a lower penalty, but the points that are far away to $\alpha_i$ will have a larger weight, which favours the zero entries in solution $c_i$.
We can use the same strategy as SMCE to solve the optimization problem in Equation \ref{eq:solve_sparseneighbor} with Alternating direction method of multipliers (ADMM) \cite{ADMM}.

As a result, we can obtain the sparse solutions $C_{P\times P}=[c_1, ..., c_P]^T$ with a few number of non-zero elements that contain the informations and connections between the projected data point and its estimated sparse neighborhoods. 
As investigated in SMCE \cite{SMCE}, in order to build the affinity matrix with sparse solution $C_{P\times P}$ we can formulate a sparse weight matrix $\Omega_{P \times P}$ with vector $\omega_i$, which is built by $\omega_{ii}=0, \omega_{ij}=\frac{c_{ij}/X_{ij}}{\sum_{t\neq i}c_{it}/X_{ti}}, j\neq i$. The achieved weight matrix $\Omega_{P\times P}$ contains only a few non-zero entries in column, which give the indices of all the estimated sparse neighbors and the distances between them.
Hence, we can collect each data $\alpha_i$ and its estimated sparse neighbors $\mathcal{N}_i$ into one local subspace $\widehat{S_i}$ according to the non-zero elements of $\omega_i$.

\subsection{Error Estimation}

\begin{figure}[!htbp]
 \centering
   \includegraphics[scale=0.3]{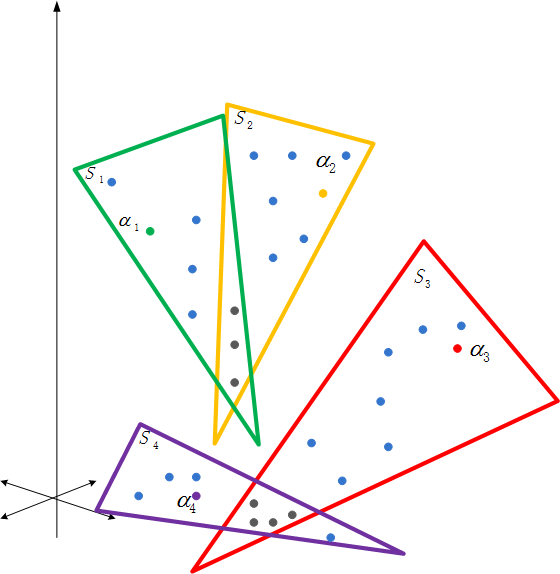}
   \caption{The geometrical illustration of incorrect local subspace estimation with sparse neighbors. $S_1, S_2, S_3, S_4$ are four estimated local subspaces spanned by the observed data $\alpha_1, \alpha_2, \alpha_3, \alpha_4$ respectively. }
   \label{fig:incorrect_estimation}
 \end{figure}

Although the sparse neighbors optimization can help us to avoid the intersection between different local subspaces, it turned into quite sensitive and can't ensure to carry all the information about the underlying local subspaces under the missing data situation.
The local subspace estimation after the sparse neighbors searching can be illustrated with Figure~\ref{fig:incorrect_estimation}. In Figure~\ref{fig:incorrect_estimation} the estimated local subspaces are not completely spanned by each observed data and its corresponding sparse neighborhood. Obviously, there are some neighbors have been estimated to span two different local subspaces, which can be called the overlapping estimation. Moreover, the obtained local subspaces with some overlapping problems cannot carry the enough dissimilarity or similarity information between two local subspaces, which can be used to build an affinity matrix that can separate the different subspaces with spectral clustering.  

For purpose of estimating these overlapping and making a strong connection between the data points from the same local subspace, we propose the following error function with Equation \ref{eq:error_estimation}
\begin{equation}
e_{it}=\|(I-\widehat{\beta_i}\widehat{\beta_i}^+)\alpha_t\|_2^2, t=1, ...,P
\label{eq:error_estimation}
\end{equation} 
where $\widehat{\beta_i}\in R^{m\times m_i}$ is the basis of estimated local subspace $\widehat{S_i}, m_i=rank(\widehat{S_i})$, which can be achieved through the SVD of $\widehat{S_i}$, and $\widehat{\beta_i}^+$ is the \textit{Moore-Penrose inverse} of $\widehat{\beta_i}$, the $I\in R^{m\times m}$ is an identity matrix. Actually the geometrical meaning of the error function $e_{it}$ is the distance between the estimated local subspace and projected data. More specifically, if the projected data $\alpha_t$ truly comes from the local subspace $\widehat{S_i}$, the corresponding error $e_{it}$ should have a very small value, which ideally is near to zero, and vice versa. 
As a consequence, after computing for each estimated local subspace $\widehat{S_i}$ its corresponding error vector $e_i=[e_i1, ..., e_iP]$, we can build an error matrix $\mathbf{e}_{P\times P}=[e_1, ..., e_P]$, which contains the strong connection between the projected data span a same local subspace.

In the end, we can construct our affinity graph $\mathcal{G}=(V, E)$ with combining the estimated error matrix $\mathbf{e}_{P\times P}$ and the sparse weight matrix $\Omega_{P\times P}$, whose the nodes $V$ represent all the projected data points and edges $E$ denote the distances between them. In our affinity graph, the connection between each two nodes $\alpha_i$ and $\alpha_j$ is determined by both the $e_{ij}$ and $\omega_{ij}$. 
Therefore, our constructed affinity graph contains only several connected elements, which are related to the data points span the same subspace, whereas there is no connection between the data points live in a different subspace. More formally, the adjacent matrix of the affinity graph is formulated as follows
\begin{equation}
\begin{aligned}
&A[i] = |\omega_i|+|{e}_i|\\
&\mathcal{A} = \left[ \begin{array}{cccc}
A[1]& 0 & ... & 0 \\
0 & A[2] & ... & 0\\
\vdots & \vdots & \ddots & \vdots\\
0 & 0 & ... & A[P]
\end{array}\right] \Gamma
\end{aligned}
\label{eq:affinity_matrix}
\end{equation}
where the $\Gamma \in R^{P\times P}$ is an arbitrary permutation matrix.
Subsequently, we can perform the normalized spectral clustering \cite{spectral_tutorial} on the symmetric matrix $\mathcal{A}$ and obtain the final clusters with different labels, and each cluster is related to one moving object.

\section{Experimental Results}
\label{sec:exp}

Our proposed framework is evaluated on both the Hopkins 155 dataset \cite{hopkins155} and the Freiburg-Berkeley Motion Segmentation Dataset~\cite{PAMI} with comparing with state-of-the-art subspace clustering and affinity-based motion segmentation algorithms. 

\textbf{Implementation Details} Most popular subspace based motion segmentation methods \cite{SSC,LSA,ALC,MSMC,PAMI} have assumed that the number of motions has been already known. For the Hopkins 155 dataset, we give the exactly number of clusters according to the number of motions, while for the Berkeley dataset we set the number of clusters with 7 for all the test sequences.   
In this work, the constrained area for searching the sparse neighbors is firstly varied in a range variables $[10, 20, 30, 50, 100]$, then it turns out that the tuned constrained area performs equally well from 20 to 50, so we choose to set the number with 20, which according to the alternative number of sparse  numbers.
In our experiments, we have applied the PCA and sparse PCA for evaluating the performance of our framework on estimating the multiple local subspaces from a general global subspace with dimension $m=5$. The sparsity controlling parameter for sparse PCA is setted to $\gamma=0.01$ and the distinct parameter vector $[\mu_1, ..., \mu_m]$ is setted to $[1/1, 1/2, ...,1/m]$. 

\subsection{The Hopkins 155 Dataset}
\begin{figure*}[htbp!]
 \centering
   \subfigure[]{
   \label{subfig:ck1_gt}
   \includegraphics[scale=0.25]{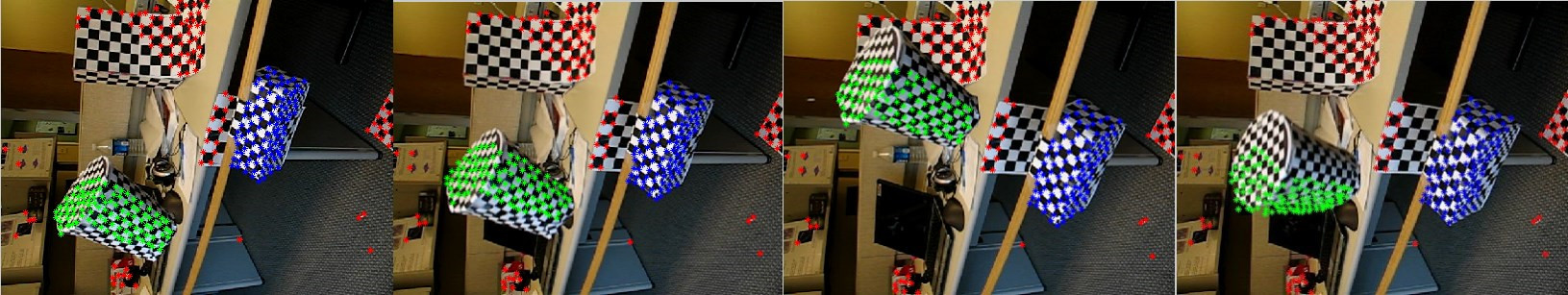}} 
   \subfigure[]{
   \label{subfig:ck1_gpca}
   \includegraphics[scale=0.25]{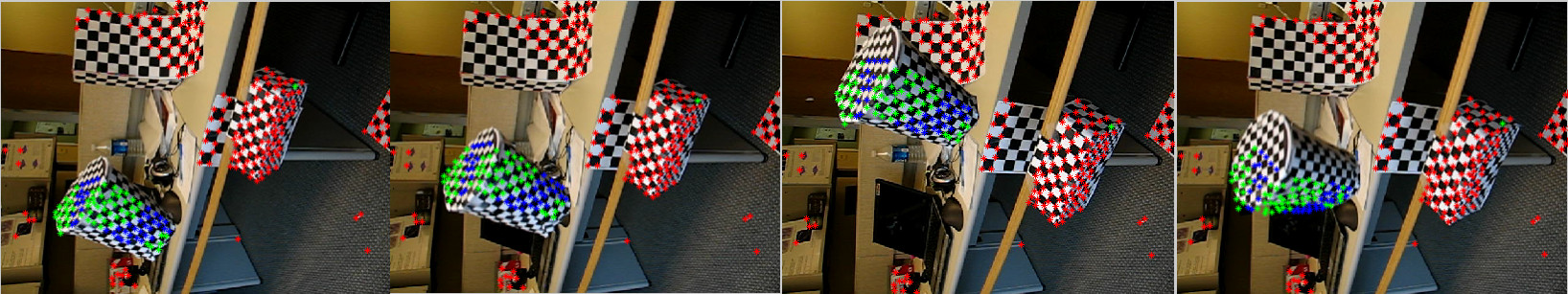}} 
   \subfigure[]{
   \label{subfig:ck1_lsa}
   \includegraphics[scale=0.25]{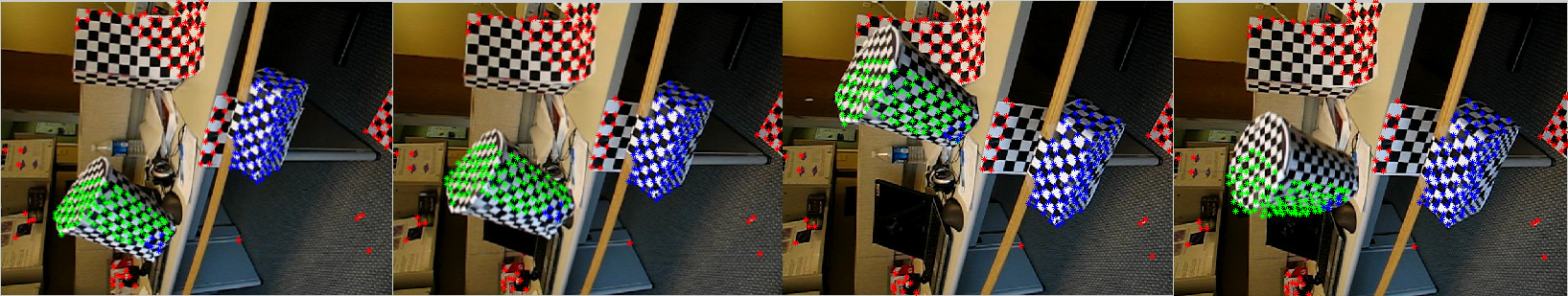}}
   \subfigure[]{
  \label{subfig:ck1_ransac}
  \includegraphics[scale=0.25]{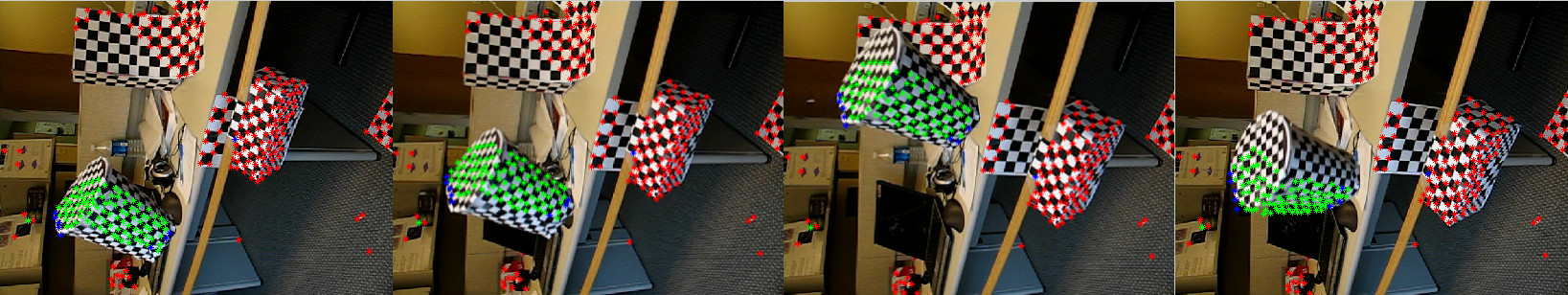}}
  \subfigure[]{
  \label{subfig:ck1_ssc}
  \includegraphics[scale=0.25]{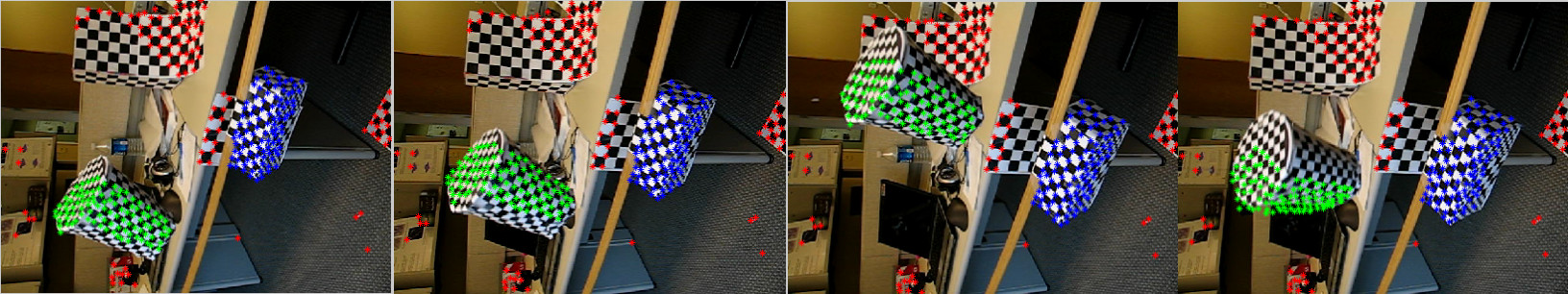}}
  \subfigure[]{
  \label{subfig:ck1_our}
  \includegraphics[scale=0.25]{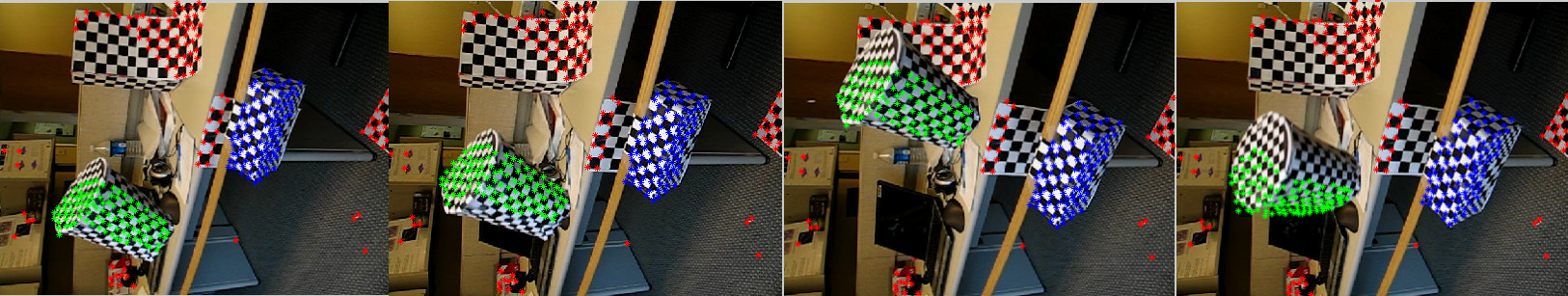}}
   \caption{Comparison of Our approach with ground truth and the other approaches on the \textit{1RT2RC} video: 
  \ref{subfig:ck1_gt}: GroudTruth; \ref{subfig:ck1_gpca}: GPCA, error: $44.98\%$; \ref{subfig:ck1_lsa}: LSA, error:$1.94\%$; \ref{subfig:ck1_ransac}: RANSAC, error: $33.66\%$; \ref{subfig:ck1_ssc}: SSC, $0\%$; \ref{subfig:ck1_our}: Our, $0\%$ on the \textit{1RT2TC} sequence from the Hopkins 155 dataset.}
\label{fig:result_ck1}
\end{figure*}

\begin{figure*}[htbp!]
 \centering
   \subfigure[]{
   \label{subfig:cars207_gt}
   \includegraphics[scale=0.25]{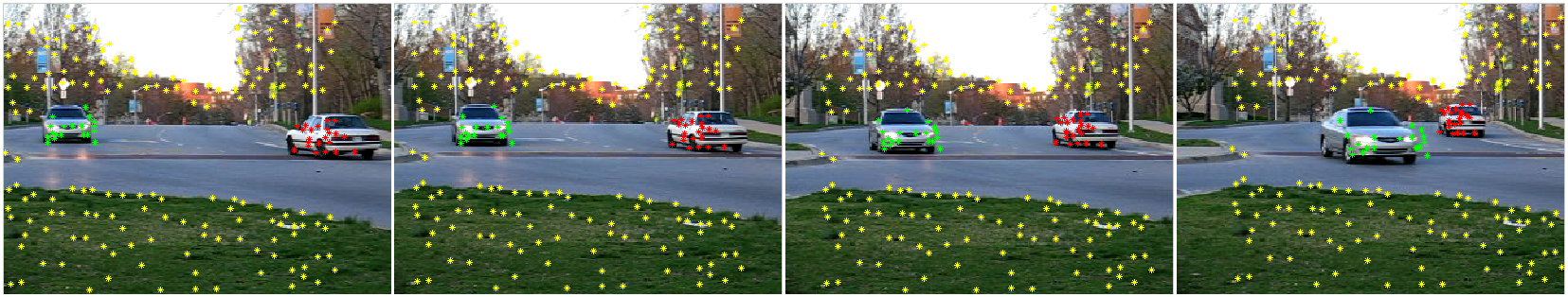}}
   \subfigure[]{
   \label{subfig:cars207_gpca}
   \includegraphics[scale=0.25]{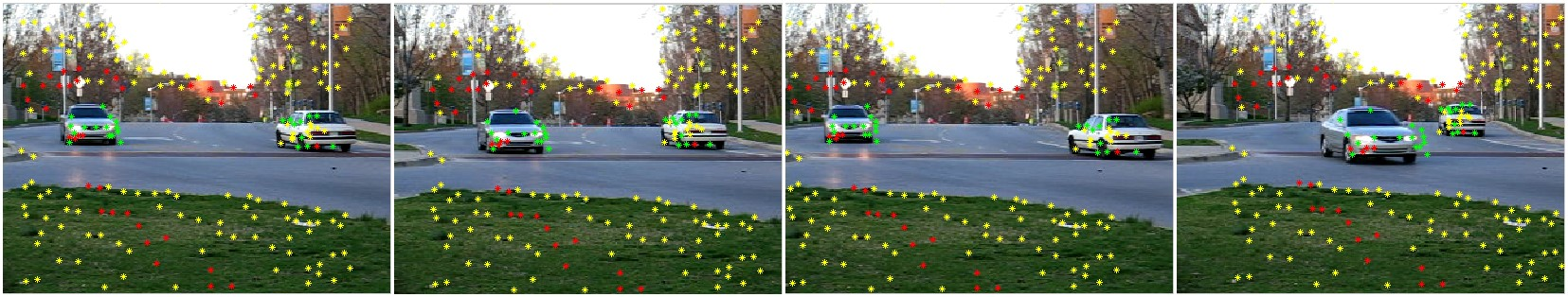}} 
   \subfigure[]{
   \label{subfig:cars207_lsa}
   \includegraphics[scale=0.25]{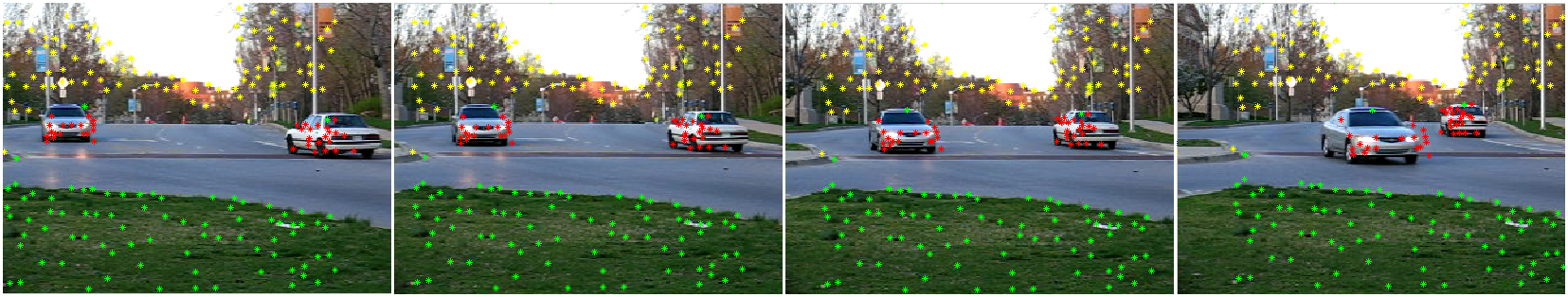}}
   \subfigure[]{
  \label{subfig:cars207_msmc}
  \includegraphics[scale=0.25]{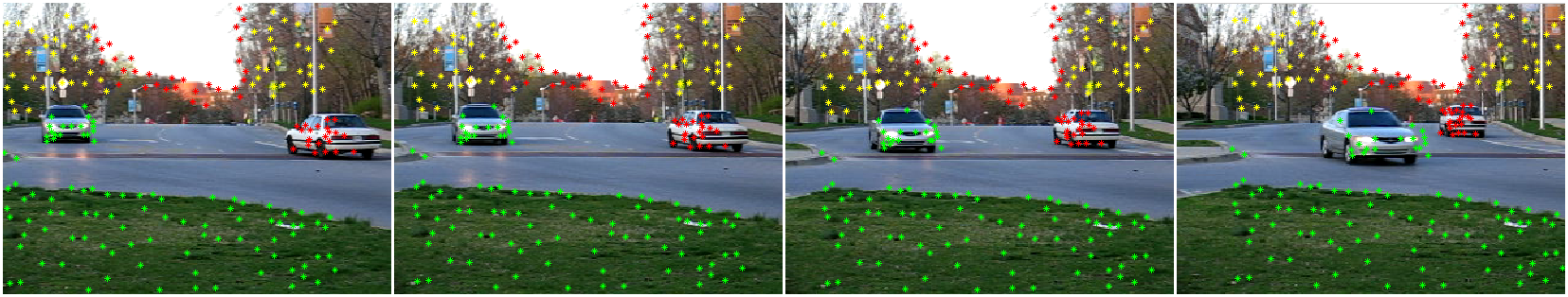}}
  \subfigure[]{
  \label{subfig:cars207_ssc}
  \includegraphics[scale=0.25]{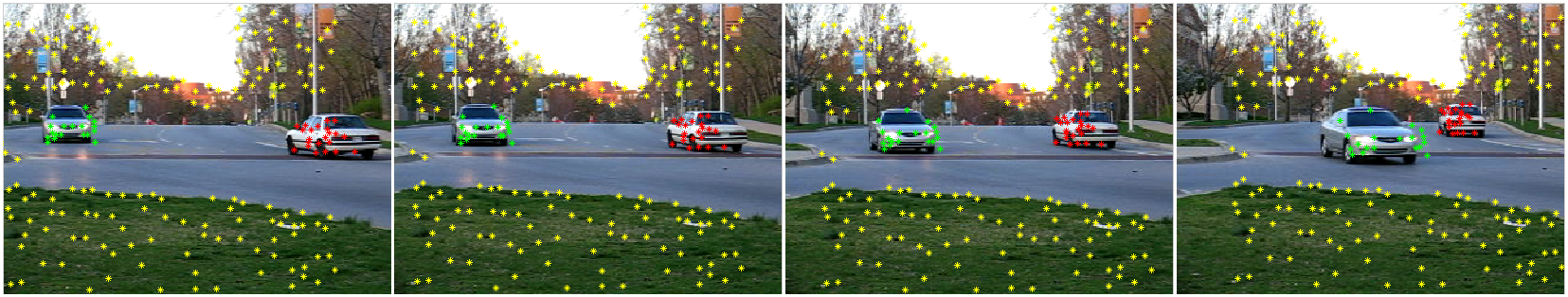}}
  \subfigure[]{
  \label{subfig:cars207_our}
  \includegraphics[scale=0.25]{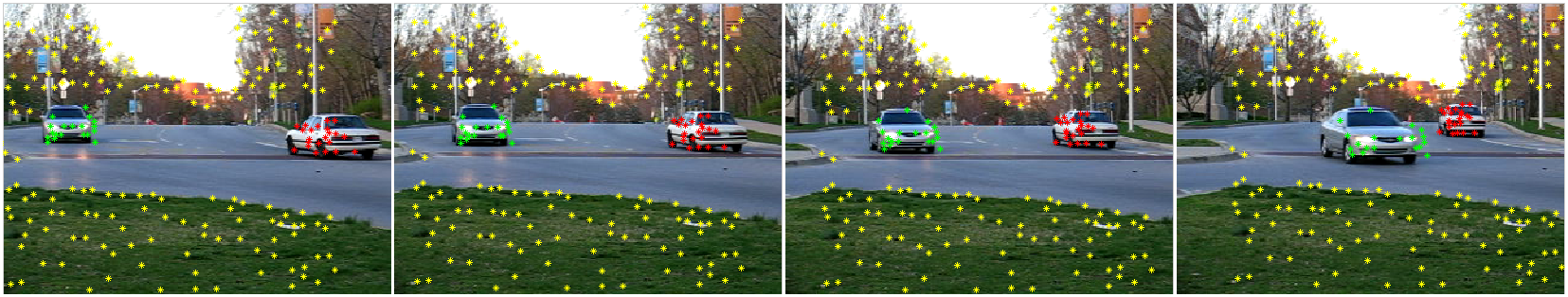}}
  \caption{Comparison of Our approach with ground truth and the other approaches on the \textit{1RT2RC} video: 
  \ref{subfig:cars207_gt}: GroudTruth; \ref{subfig:cars207_gpca}: GPCA, error: $19.34\%$; \ref{subfig:cars207_lsa}: LSA, error:$46.23\%$; \ref{subfig:cars207_msmc} MSMC, error: $46.23\%$; \ref{subfig:cars207_ssc} SSC, $0\%$; \ref{subfig:cars207_our}: Our, $0\%$.}
\label{fig:result_traf}
\end{figure*}

The Hopkins 155 dataset \cite{hopkins155} contains 3 different kinds sequences: checkerboard, traffic and articulated. For each of them, the tracked feature trajectories are already been provided in the ground truth and the missing features are removed as well, which means the trajectories in the Hopkins 155 dataset are fully observed and there is no missing data. 
We have computed the average and median misclassification error for comparison our method with state-of-the-art methods: SSC \cite{SSC}, LSA \cite{LSA}, ALC \cite{ALC}and MSMC \cite{MSMC}, as shown in Table \ref{table:2motionsHopkins}, Table \ref{table:3motionsHopkins}, Table \ref{table:allmotionsHopkins}. 
Table \ref{table:computationtimes} refers to the run times of our method comparing with two sparse optimization based methods: ALC and SSC.
\begin{table}[h]\small
\begin{center}
\begin{tabular}{|l|c|c|c|c|c|c|}
\hline
Method & ALC & SSC & MSMC & LSA  & Our$_{pca}$ & Our$_{spca}$ \\
\hline\hline
  \multicolumn{7}{|l|}{Articulated, 11 sequences} \\
   \cline{1-7} 
   mean & 10.70 & 0.62 & 2.38 & 4.10 & 2.67 & \textbf{0.55}\\
    median & 0.95 & 0.00 & 0.00 & 0.00 & 0.00 & 0.00\\
 \hline
 \multicolumn{7}{|l|}{Traffic, 31 sequences} \\
 \cline{1-7}
   mean & 1.59 & \textbf{0.02} & 0.06 & 5.43 & 0.2 & 0.48 \\
   median & 1.17 & 0.00 & 0.00 & 1.48 & 0.00 & 0.00 \\
 \hline
  \multicolumn{7}{|l|}{ Checkerboard, 78 sequences}\\
 \cline{1-7}
   mean& 1.55 & 1.12 & 3.62 & 2.57 & 1.69 & \textbf{0.56} \\
   median & 0.29 & 0.00 & 0.00 & 0.27 & 0.00 & 0.00 \\
   \hline
\multicolumn{7}{|l|}{ All 120 sequences} \\   
\cline{1-7}   
   mean & 2.40 & 0.82 & 2.62 & 3.45 & 1.52 & \textbf{0.53}\\
   median & 0.43 & 0.00 & 0.00 & 0.59 & 0.00 & 0.00\\
   \hline
\end{tabular}
\end{center}
\caption{Mean and median of the misclassification (\%) on the Hopkins 155 dataset with 2 motions.}
\label{table:2motionsHopkins}
\end{table}
\begin{table}[h]\small
\begin{center}
\begin{tabular}{|l|c|c|c|c|c|c|}
\hline
Method & ALC & SSC & MSMC & LSA  & Our$_{pca}$ & Our$_{spca}$ \\
\hline\hline
  \multicolumn{7}{|l|}{Articulated, 2 sequences} \\
   \cline{1-7} 
    mean & 21.08 & 1.91 & \textbf{1.42} & 7.25 & 3.72 & 3.19\\
    median & 21.08 & 1.91 & 1.42 & 7.25 & 3.72 & 3.19\\
 \hline
 \multicolumn{7}{|l|}{Traffic, 7 sequences} \\
 \cline{1-7}
   mean & 7.75 & 0.58 & \textbf{0.16} & 25.07 & 0.19 & 0.72\\
   median & 0.49 & 0.00 & 0.00 & 5.47 & 0.00 & 0.19\\
 \hline
  \multicolumn{7}{|l|}{ Checkerboard, 26 sequences}\\
 \cline{1-7}
  mean & 5.20 & 2.97 & 8.30 & 5.80 & 5.01 & \textbf{1.22}\\
   median & 0.67 & 0.27 & 0.93 &1.77 & 0.78 & 0.55\\
   \hline
\multicolumn{7}{|l|}{ All 35 sequences} \\   
\cline{1-7}   
  mean & 6.69 & 2.45 & 3.29 & 9.73 & 2.97 & \textbf{1.94}\\
   median & 0.67 & 0.20 & 0.78 & 2.33 & 1.50 & 1.30\\
   \hline
\end{tabular}
\end{center}
\caption{Mean and median of the misclassification (\%) on the Hopkins 155 dataset with 3 motions.}
\label{table:3motionsHopkins}
\end{table}
\begin{table}[h]\small
\begin{center}
\begin{tabular}{|l|c|c|c|c|c|c|}
\hline
Method & ALC & SSC & MSMC & LSA  & Our$_{pca}$ & Our$_{spca}$ \\
\hline\hline
  \multicolumn{7}{|l|}{all 155 sequences} \\
   \cline{1-7} 
   Mean  & 3.56 & 1.24 & 2.96 & 4.94 & 1.98 & \textbf{0.70}\\
   Median  & 0.50 & 0.00 &  & 0.90 & 0.75 & 0.00\\
 \hline
\end{tabular}
\end{center}
\caption{Mean and median of the misclassification (\%) on all the Hopkins 155 dataset.}
\label{table:allmotionsHopkins}
\end{table}
\begin{table}[h]\small
\begin{center}
\begin{tabular}{|l|c| l | l | l |}
\hline
Method & ALC & SSC & Our$_{PCA}$ & Our$_{SPCA}$ \\
\hline\hline
 Run-time [s] & 88831 & 14500 & \textbf{1066} & \textbf{1394}\\
 \hline
\end{tabular}
\end{center}
\caption{Computation-Time (s) on all the Hopkins 155 dataset.}
\label{table:computationtimes}
\end{table}
Obviously, as Table \ref{table:2motionsHopkins} and Table \ref{table:3motionsHopkins} show, the overall error rate of ours with sparse PCA projection is the lowest for both 2 and 3 motions. 
Generally, the PCA projection has a lower accuracy than sparse PCA projection for the articulated and checkerboard sequences. 
However, the traffic video with PCA projection reaches a better result than the sparse PCA projection, which gives us a conclusion that PCA is more robust to represent the rigid motion trajectory matrix, but the sparse PCA projection can better represent the trajectory matrix of independent or non-rigid motions.
We also notice that MSMC performs the best for the traffic sequence with 3 motions, but our work with PCA projection is just slightly worse to MSMC and inferior to SSC, which is one of the most accurate subspace-based algorithm. 
But due to the property of MSMC, which is based on computing the affinities between each pair trajectories, it is highly time-consuming.
The checkerboard data is the most significant component for the entire Hopkins dataset, which in particular contains a lot of features points and many intersection problems between different motions.
To be specific, the most accurate results for the checkerboard sequences belong to our proposed framework with sparse PCA projection, either for two or three motions. It means that our method has the most accuracy for clustering different intersected motions.
Table \ref{table:allmotionsHopkins} shows our method achieves the least misclassification error for all the sequences from the Hopkins dataset in comparison with all the other algorithms. 
Although our method with sparse PCA or PCA projection is a bit loss of precision for the traffic sequences, we save a lot of computation times comparing with SSC and ALC as shown in Table \ref{table:computationtimes}.
We evaluate our method with sparse PCA projection in comparison with LSA \cite{LSA}, SSC \cite{SSC}, MSMC \cite{MSMC}, GPCA \cite{GPCA}, RANSAC \cite{ransac} and MSMC \cite{MSMC} in Figure~\ref{fig:result_ck1} and Figure~\ref{fig:result_traf} on the Hopkins 155 dataset. 
Note that MSMC has not been evaluated on the checkboard sequence.

\subsection{Freiburg-Berkeley Motion Segmentation Dataset}
In this subsection, our method has been evaluated on the Freiburg-Berkeley Motion Segmentation dataset \cite{PAMI} to test the performance on the real video sequences with occlusion and moving camera problems. This dataset contains 59 sequences and all the feature trajectories are tracked densely. All the missing trajectories have not been removed and there is no pre-processing for correcting the error tracked trajectory. The parameters for evaluation are precision (\%) and recall (\%). 
Our method has been compared with Ochs \cite{PAMI}, which is based on the affinity of the trajectories between each two frames, SSC \cite{SSC} and ALC \cite{ALC}. The results on all the training set and test set of the Berkeley dataset are shown in Table \ref{table:berkeleyResults}.
\begin{table}[h]\small
\begin{center}
\label{table:berkeleyResults}
\begin{tabular}{|l|c|c|c|c|c|}
   \hline
     & Ochs & ALC & SSC & Our$_{pca}$ & Our$_{spca}$  \\
   \hline 
   Precision & \textbf{82.36} & 55.78 & 64.55 & 72.12 & 70.77 \\
   \hline
   Recall & 61.66 & 37.43 & 33.45 & \textbf{66.52} & 65.42 \\
   \hline
\end{tabular}
\end{center}
\caption{Results on the entire Freiburg-Berkeley Motion Segmentation Dataset~\cite{PAMI}. }
\end{table}   

\begin{figure*}[!htbp]
\begin{center}
\subfigure[]{
\includegraphics[width=0.95\textwidth]{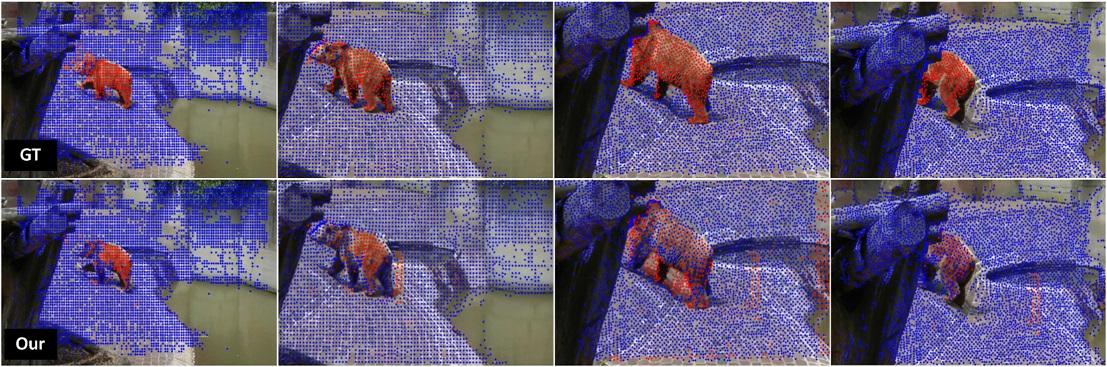}
\label{subfig:berkeley_bear}}
\subfigure[]{
\includegraphics[width=0.95\textwidth]{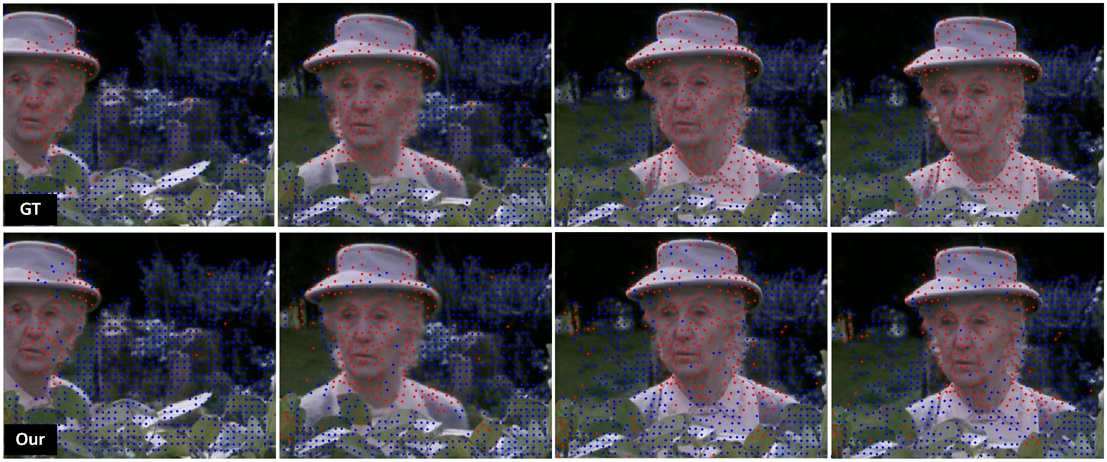}
\label{subfig:berkeley_marple4}}
\subfigure[]{
\includegraphics[width=0.95\textwidth]{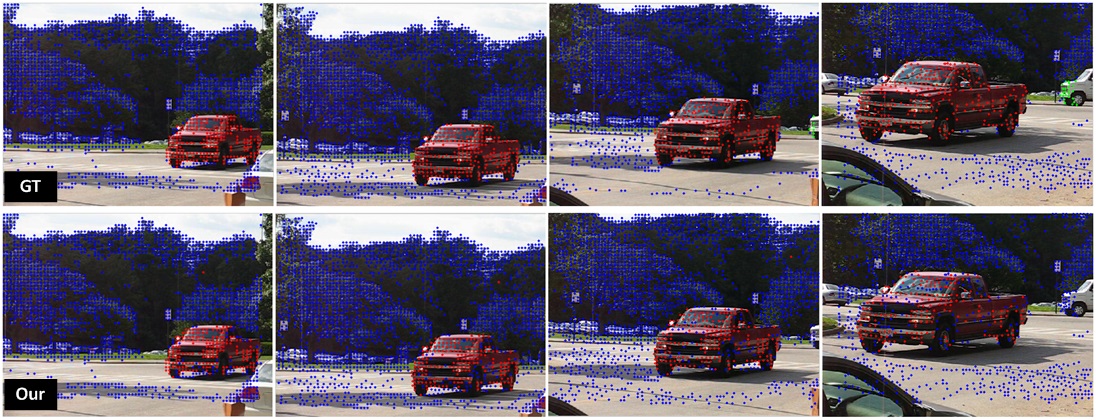}
\label{subfig:berkeley_cars8}}
\end{center}
   \caption{Our segmentation results on Freiburg-Berkeley Motion Segmentation Dataset in comparison with the groundtruth segmentations from \cite{PAMI}. \ref{subfig:berkeley_bear}:bear01, \ref{subfig:berkeley_marple4}: marple4, \ref{subfig:berkeley_cars8}: cars8.}
\label{fig:results_Berkeley}
\end{figure*}
%
\begin{figure*}[!htbp]
\begin{center}
\includegraphics[width=0.95\textwidth]{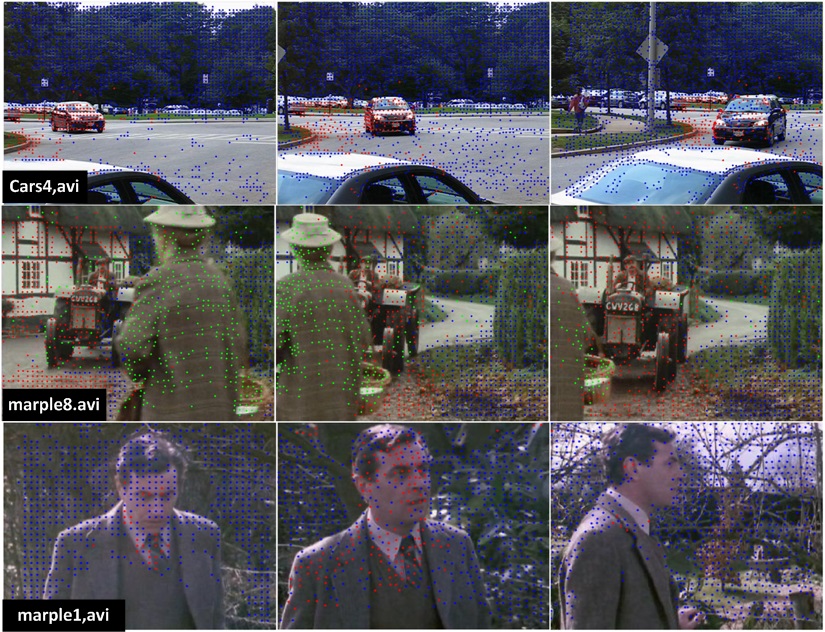}
\end{center}
   \caption{Additional segmentation results of Freiburg-Berkeley Motion Segmentation Dataset \cite{PAMI}.}
\label{fig:results_Berkeleyadd}
\end{figure*}

In general, as shown in Table \ref{table:berkeleyResults}, the PCA projection has a better performance on this dataset than the sparse PCA, which can not deal with the data matrix contains a lot of zero entries.
More specifically, our method with PCA projection obtains the most Recall value comparing with the others, which indicates our assigned clusters can cover the most parts of the different ground-truth regions.
However, compared with Ochs \cite{PAMI}, which is based on the affinity, our method lacks the precision. 
It means that our method can detect the boundaries of different regions but can not complete segment the moving objects from the background. 
Figure \ref{fig:results_Berkeley} show us the examples of our results with PCA projection. Among all of these examples, our method has high quality segmentations of the primary foreground moving objects, which according to to the high recall value. However, there are some incorrect segmentations as well, such as the features on the object cannot be distinguished exactly especially at the last few frames. These incomplete segmentation results indicate the small precision value in Table \ref{table:berkeleyResults}. 
Among all of the subspace-based motion segmentation algorithms SSC and ALC, which need to firstly apply the sparse reconstruction for the incomplete trajectories, our method only depends on the error estimation and sparse neighbors technique but has a superior performance on the precision and recall.

Figure \ref{fig:results_Berkeleyadd} show us some additional segmentation results. 
The typical failure segmentations are shown in the bottom row \textit{marple1.avi}, which contains 300 frames. Our method can not exactly extract the moving objects from the background for the video that has the really long observed frames. Moreover our method can not segment the video accurately when the camera is also moving, due to the moving foreground usually has the short feature trajectories that are very difficult to handle.

\section{Conclusions}
\label{sec:con}

In this paper, we have proposed a subspace-based framework for segmenting multiple moving objects from a video sequence with integrating global and local sparse subspace optimization methods. 
The sparse PCA performs a data projection from a high-dimensional subspace to a global subspace with sparse orthogonal principal vectors. 
To avoid improperly choosing K-nearest neighbors and defend intersection between different local subspaces, we seek a sparse representation for the nearest neighbors in the global subspace for each data point that span a same local subspace. 
Moreover, we propose an error estimation to refine the local subspace estimation for the missing data. 
The advantage of the proposed method is that we can apply two sparse optimizations and a simple error estimation to handle the incorrect local subspace estimation under the missing trajectories.
The limitation of our work is the number of motions should be known firstly and only a constrained number of missing data can be handled accurately. The experiments on the Hopkins and Berkeley dataset show our method are comparable with state-of-the-art methods in terms of accuracy, and sometimes exceeds them on both precision and computation time.

\section*{Acknowledgements}
\label{ACKNOWLEDGEMENTS}

The work is funded by DFG (German Research Foundation) YA 351/2-1 and the ERC-Starting Grant (DYNAMIC
MINVIP).
The authors gratefully acknowledge the support.



\end{document}